\documentclass[conference]{IEEEtran}
\IEEEoverridecommandlockouts
\usepackage{multirow}
\usepackage{cite}
\usepackage{amsmath,amssymb,amsfonts}
\usepackage{algorithmic}
\usepackage{graphicx}
\usepackage{textcomp}
\usepackage{xcolor}
\usepackage{booktabs}
\usepackage{float}
\usepackage{subcaption}
\def\BibTeX{{\rm B\kern-.05em{\sc i\kern-.025em b}\kern-.08em
    T\kern-.1667em\lower.7ex\hbox{E}\kern-.125emX}}
\begin{document}

\title{Transformer-Based Wildlife Species Classification from Daily Movement Trajectories\\
}

\author{\IEEEauthorblockN{Obed Irakoze}
\IEEEauthorblockA{\textit{Department of Electrical \& Computer Engineering} \\
\textit{Carnegie Mellon University Africa}\\
Kigali, Rwanda \\
oirakoze@andrew.cmu.edu}
\and
\IEEEauthorblockN{Prasenjit Mitra}
\IEEEauthorblockA{\textit{Department of Electrical \& Computer Engineering} \\
\textit{Carnegie Mellon University Africa}\\
Kigali, Rwanda \\
prasenjm@andrew.cmu.edu}

}

\maketitle

\begin{abstract}

Inferring the identity of wildlife species from daily movement data alone is a challenging task. 
We train sequence models on large-scale, 7-species GPS trajectories from the Movebank platform. Trajectories models are evaluated using a protocol in which entire telemetry studies or regions are held out during testing. We compare Transformer-based sequence models to LSTM, CNN, and Temporal Convolutional Networks, and find that Transformers consistently achieve higher balanced accuracy with gains of approximately 8 to 22 percentage points, depending on the species and experimental setting. In an elephant binary classification task with 1-hour resolution, the Transformer achieves a balanced accuracy of 0.83 and an AUC of 0.92, substantially outperforming all baseline models. We examine, under data-limited conditions, feature representations by analyzing the differences between a basic displacement-based encoding and an expanded range of movement descriptors that include speed, direction, and turning behavior. With feature augmentation, we see clear performance gains, especially for underrepresented and sparsely represented species, such as large carnivores, lions, and Zebras. Finally, experiments comparing 1-hour and 30-minute temporal resolutions show that while finer sampling can capture short-term movement patterns for some species, a unified 1-hour resolution yields more promising performance across studies by reducing missing data and ensuring consistent temporal coverage.

\end{abstract}

\begin{IEEEkeywords}
Wildlife trajectory analysis, Species classification, Sequence modeling.
\end{IEEEkeywords}

\section{Introduction}

Advances in animal-borne telemetry and open data portals such as Movebank have facilitated the accumulation of wildlife movement trajectories between species, locations, and studies\cite{kays2015terrestrial}. 
We address whether the identity of species of wildlife can be inferred solely from movement trajectories, without relying on absolute geographic location, habitat descriptors, or environmental covariates. Here, a study is defined as an independent telemetry effort in a specific area at the same time with its own sampling method, collar technology, and environmental context\cite{hebblewhite2010distinguishing}. Models trained on data from a single study or region can achieve strong performance locally, but they often fail when applied to other areas, as movement trajectories are shaped not only by species behavior, but also by region-specific factors such as fences, roads, land-use patterns and human activity\cite{valavi2019blockcv}. The development of multiple separate models for each study or region is often not
desirable, as there are no labeled data everywhere and new monitoring locations continue to be added. 
Combining data across regions allows us to assess whether the same species exhibits consistent movement patterns across different parks and countries despite local differences\cite{nathan2008movement}. 
We seek to develop models that generalize across studies and regions, allowing species classification to rely on intrinsic movement behavior rather than site-specific routine and idiosyncrasies, and to remain usable in areas without data or previously unseen \cite{roberts2017cross}\cite{yates2018outstanding}.

Our objective is to build a model that predicts the entity of species using movement trajectories alone, and our dataset comprises 7 species. In contrast to using contextual or environmental variables, we explicitly examine whether species can be distinguished from each other based only on the movement of the species. In addition to the prediction accuracy, we study the movement behaviors that enable successful classification of the species in order to describe movement behavior patterns that differ between the species on a daily basis. In terms of movement, this can include the structure of movement in space, the speed of movement, the direction of movement, the turning angles, or the patterns of movement. Using telemetry data from varying conditions in the region, we analyze species-specific movement patterns and how these patterns can be used to identify the species.

There are several factors that make this problem particularly challenging. One of them is that animal telemetry data exhibit strong spatial, temporal, and individual-level autocorrelation. Given that  consecutive observations of trajectories from the same animal or region are highly similar, random train–test splits can inflate the performance of the model by allowing models to learn where an animal is rather than how it moves \cite{roberts2017cross}. Secondly, telemetry datasets differ widely in their temporal sampling resolution across studies. Resampling movement trajectories to a large temporal resolution intervals can suppress short-term movement dynamics like pauses, directional changes, or bursts of activity that are often ecologically informative and they potential species pattern that might distinguish the species\cite{he2023guide}. Therefore, balancing temporal resolution consistency across studies which preserve informative movement patterns, presents a fundamental challenge for trajectory-based species classification.

Most prior work on animal movement modeling has focused on behavioral state classification~\cite{wang2019machine}, habitat selection~\cite{nams2014combining}, or within-study prediction, often using Hidden Markov Models, step-selection functions, or recurrent neural networks evaluated under individual-level or random cross-validation schemes\cite{patterson2008state}. Although recent studies emphasize spatially structured validation and the risks of location bias\cite{roberts2017cross}, few explicitly address cross-study species classification from movement alone, and fewer still examine how feature representation and temporal resolution interact with modern attention-based sequence models. Consequently, it remains unclear whether the performance gains reported reflect true species-level movement signatures or artifacts of study-specific sampling and geography\cite{valavi2019blockcv, roberts2017cross}.

In this work, we study the classification of wildlife species from movement trajectories by learning species-specific 
movement patterns. We represent each animal’s movement as an ordered sequence of GPS observations collected on a single UTC calendar day. For each animal-day, this yields sequences of up to 24 positions at 1-hour temporal resolution or 48 positions at 30-minute resolution, which
are used as input to sequence models trained to predict species identity. 

We focus on African wildlife species that are well represented in open-access telemetry datasets and exhibit diverse movement behaviors within a shared continental context. This allows us to train models on data collected in some regions and test them on data from different regions, evaluating whether species can be distinguished based on movement patterns rather than location-specific characteristics. Using Movebank data\cite{kays2022movebank}, we systematically select seven African species with open-access high-resolution trajectories available for download and analysis—baboon, buffalo, caracal, Zebra, elephant, lion and wildebeest. Individual Movebank telemetry studies may include multiple species, and species are represented by varying numbers of tracked animals. 
We compare a Transformer-based sequence model\cite{vaswani2017attention} with baselines from the LSTM, CNN, and Temporal Convolutional Network \cite{hochreiter1997long}. In studies held-out for the test dataset, the Transformer achieves a balanced accuracy of 0.81 and an AUC of 0.92, outperforming baselines whose balanced accuracy ranges from 0.68 to 0.77 and whose AUC ranges from 0.78 to 0.87. Moreover, enhancing a minimal displacement-based encoding with additional movement descriptors derived from the same trajectories -- speed, bearing, and turning angle---improves balanced accuracy by 43.10\%. Finally, we evaluated the effect of temporal resolution and showed that although finer 30-minute resampling can reflect short-term movement dynamics for certain species, a common 1-hour resolution is better suited for cross-study modeling due to reduced missingness and more consistent temporal coverage\cite{patterson2017statistical}.

\section{Related Work}

\subsection{Traditional Movement Modeling and Species Identification}
The analysis of animal movement trajectories is grounded in the movement ecology paradigm \cite{nathan2008movement}. Traditional approaches rely on correlated random walks, step-selection functions, and Hidden Markov Models (HMMs) to infer latent behavioral states (e.g., foraging, resting) from GPS telemetry \cite{patterson2017statistical,avgar2016integrated}. While highly effective for behavioral segmentation, these generative models are typically species- or region-specific and are not designed for species-level classification.

Species identification from telemetry data remains comparatively underexplored. Existing studies predominantly employ classical machine learning models, such as Random Forests, relying heavily on hand-crafted features \cite{wang2019machine}. Crucially, prior evaluations often utilize random train-test splits. As highlighted in \cite{roberts2017cross} and \cite{valavi2019blockcv}, this practice introduces geographic leakage, allowing models to memorize site-specific environmental routines rather than intrinsic, generalizable species behaviors. Consequently, models trained in this manner can lead to overly
optimistic performance and limited generalization to unseen regions or studies\cite{meyer2022machine}.

\subsection{Deep Learning for Trajectory Classification}
Deep learning (DL) facilitates the end-to-end modeling of movement trajectories as multivariate time series \cite{chen2024deep}. Sequential architectures like Recurrent Neural Networks (RNNs), LSTMs, and Temporal Convolutional Networks (TCNs) have shown promise for behavioral state inference and short-term trajectory prediction \cite{ren2020st}. Recently, Transformers \cite{vaswani2017attention} have emerged as powerful sequence encoders, adept at modeling long-range temporal dependencies in daily movement sequences. 

However, the application of DL to identify species across distinct geographical regions remains nascent. Most existing DL studies evaluate performance within a single dataset, leaving their capacity for multi-study generalization untested. Our study directly addresses this gap. By modeling daily trajectories using Transformers and explicitly enforcing study-level holdouts during evaluation, our methodology aligns with recent calls for rigorous validation in AI-driven conservation \cite{tuia2022perspectives}, ensuring the extraction of true species-specific kinematic signatures rather than localized spatial artifacts.

\section{Data and Preprocessing}

This section describes the telemetry datasets used in this study and the preprocessing steps applied to construct comparable daily movement trajectories collected under different sampling protocols.

\subsection{Data Source and Species Selection}

We use GPS telemetry data sourced from Movebank, a worldwide archive of animal movement data~\cite{kays2015terrestrial}\cite{kays2022movebank}. The dataset is made up of tracking data from seven species of African wildlife: \emph{baboon}, \emph{buffalo}, \emph{caracal}, \emph{Zebra}, \emph{elephant}, \emph{lion}, and \emph{wildebeest}, and spans diverse regions of Africa and separate telemetry studies.

A study is defined here as an independent telemetry campaign conducted in a specific region and time period, typically characterized by its own sampling protocol, collar technology, and tracking duration. Individual studies may include multiple species, and species are represented by varying numbers of tracked animals. Because data originate from different studies with distinct collection protocols, models trained and evaluated using random splits risk capturing study- or location-specific patterns rather than species-level movement characteristics. 
To mitigate this effect, we evaluated models by holding out one entire study during testing for each species, ensuring that test data come from regions and data collection settings not observed during training and validation. In total, this research utilizes 16 distinct Movebank datasets purposefully selected to cover diverse geographical regions. These individual telemetry campaigns were conducted at various times between 1998 and 2023. Details of the specific Movebank studies and dataset splits used for these experiments are summarized in Table~\ref{tab:movebank_details}.

\begin{table}[!t]
\centering
\small
\caption{Movebank studies used and dataset splits.}
\label{tab:movebank_details}
\begin{tabular}{l l l l}
\toprule
Species & Movebank Study ID & Split Usage \\
\midrule
Baboon & 2131q5 (DOI) & Test \\
Baboon & 1723547 & Train/Val \\
\midrule
Lion & 220229 & Train/Val \\
Lion & 220229 & Test \\
Lion & 150531 & Train/Val \\
\midrule
Wildebeest & 132915 & Test \\
Wildebeest & 225301 & Train/Val \\
Wildebeest & 1310113 & Train/Val \\
\midrule
Elephant & 736029750 & Train/Val \\
Elephant & 1818825 & Test \\
Elephant & 3nj3qj45 (DOI) & Train/Val \\
Elephant & 1630/2970/5990 & Train/Val \\
\midrule
Buffalo & 2138 & Train/Val \\
Buffalo & 1803741 & Train/Val \\
\midrule
Caracal & 1.317 (DOI) & Train/Val/Test \\
\midrule
Buffalo/Zebra & 259966228 & Buffalo: Test \\&& Zebra: All splits \\
\bottomrule
\end{tabular}
\end{table}

\subsection{Temporal Standardization and Resampling}

Telemetry data exhibit varying sampling intervals, ranging from less than one hour to more than one hour between consecutive observations. To ensure temporal 
comparability across studies, all trajectories were resampled onto fixed temporal 
grids using two resolutions: 1-hour and 30-minute intervals. A uniform temporal 
resolution is a common prerequisite for comparative movement analysis and sequence 
modeling \cite{he2023guide}.

Let $\{(t_i, \mathbf{p}_i)\}_{i=1}^N$ denote the original GPS trajectory of an 
individual animal, where $t_i$ is the timestamp and $\mathbf{p}_i = 
(\text{lat}_i, \text{lon}_i)$ the recorded position. Prior to resampling, any raw 
observations recorded at the exact same timestamp are spatially averaged to reduce 
measurement noise while preserving temporal continuity.

We construct trajectories on a regular temporal grid with resolution $\Delta t$ 
($\Delta t = 1$~hour for hourly resampling and $\Delta t = 30$~minutes for 
half-hourly resampling). Each original timestamp $t_i$ is mapped to the nearest 
grid time via $\tilde{t}_i = \operatorname{round}_{\Delta t}(t_i)$,
where $\operatorname{round}_{\Delta t}(\cdot)$ denotes rounding to the nearest 
multiple of $\Delta t$, ensuring that the temporal displacement satisfies 
$|t_i - \tilde{t}_i| \le \Delta t/2$. If multiple observations map to the same 
grid time, only the observation with minimal $|t_i - \tilde{t}_i|$ is retained.

Let $\mathcal{T} = \{\tau_0, \tau_1, \dots, \tau_n\}$ denote the complete regular 
grid spanning from the earliest to the latest retained observation, with spacing 
$\Delta t$, and let $\mathcal{T}^* \subseteq \mathcal{T}$ denote the subset of 
grid times at which a valid (non-interpolated) observation exists. The resampled 
position $\mathbf{p}(\tau)$ at any target grid time $\tau \in \mathcal{T}$ is 
defined as:

{\small
\begin{equation}
    \mathbf{p}(\tau) =
    \begin{cases}
        \mathbf{p}_k,
            & \tau = \tau_k \in \mathcal{T}^*, \\[6pt]
        \mathbf{p}_k + \frac{\tau - \tau_k}{\tau_{k+1} - \tau_k}
            \,(\mathbf{p}_{k+1} - \mathbf{p}_k),
            & \tau_k,\tau_{k+1} \in \mathcal{T}^*, \\
            & \tau_k < \tau < \tau_{k+1}, \\
            & \tau_{k+1} - \tau_k = 2\Delta t, \\[8pt]
        \text{undefined},
            & \text{otherwise.}
    \end{cases}
\end{equation}
}
Linear interpolation is therefore applied only when exactly one intermediate grid 
point is missing between two consecutive valid observations in $\mathcal{T}^*$. 
If the temporal gap between consecutive valid observations exceeds $2\Delta t$, 
the intermediate positions remain undefined and are excluded from further analysis.

Importantly, no forward-filling or backward-filling of positions is performed. 
Repeated positions at consecutive grid times occur only when identical coordinates 
are present in the original data.

\subsection{Daily Trajectory Construction}



To prevent temporal information leakage, trajectories are segmented into daily sequences based on UTC (max 24 points for 1-hour resolution, 48 points for 30-minute). We retain days with at least 12 observations (1-hour) or 25 observations (30-minute). This coverage-based filtering balances temporal completeness with data retention \cite{valavi2019blockcv}. 

Table~\ref{tab:post_preprocessing_stats} details the final dataset composition, illustrating the retained animal-days, unique individuals, and total GPS points per species.

\begin{table}[!t]
\centering
\caption{Dataset statistics after preprocessing for 30-minute and 1-hour resampling protocols.}
\label{tab:post_preprocessing_stats}
\resizebox{\columnwidth}{!}{%
\begin{tabular}{lc|ccc|ccc}
\toprule
& & \multicolumn{3}{c|}{\textbf{30-Minute Resampling}} & \multicolumn{3}{c}{\textbf{1-Hour Resampling}} \\
\cmidrule(lr){3-5} \cmidrule(lr){6-8}
\textbf{Species} & \textbf{Studies} & \textbf{Valid Days} & \textbf{Animals} & \textbf{Points} & \textbf{Valid Days} & \textbf{Animals} & \textbf{Points} \\
\midrule
Baboon     & 2 & 65    & 38 & 21,138    & 108   & 38 & 19,406 \\
Buffalo    & 3 & 529   & 11 & 72,526    & 583   & 37 & 38,550 \\
Caracal    & 1 & 842   & 27 & 149,094   & 844   & 27 & 91,308 \\
Elephant   & 4 & 4,628 & 44 & 1,019,480 & 5,405 & 61 & 597,171 \\
Lion       & 3 & 1,126 & 27 & 489,997   & 1,448 & 29 & 282,350 \\
Wildebeest & 3 & 1,972 & 49 & 838,763   & 2,519 & 50 & 569,143 \\
Zebra      & 1 & 499   & 9  & 135,237   & 500   & 9  & 67,894 \\
\bottomrule
\end{tabular}%
}
\end{table}

\subsection{Kinematic and Temporal Encoding}
To avoid reliance on absolute geographic location, latitude ($\phi$) and longitude ($\lambda$) in radians are projected onto the unit sphere: $x = \cos\phi\cos\lambda$, $y = \cos\phi\sin\lambda$, $z = \sin\phi$. We compute instantaneous displacement vectors as $\Delta x_t = x_t - x_{t-1}$, $\Delta y_t = y_t - y_{t-1}$, and $\Delta z_t = z_t - z_{t-1}$. While unit-sphere displacement scales can technically vary by latitude, our strict study-level holdout strategy guarantees that models cannot exploit location-specific displacement magnitudes, forcing the architecture to learn generalized, relative movement motifs.

We encode time cyclically to capture circadian rhythms without boundary discontinuities \cite{kazemi2019time2vec}. For 1-hour resolution, the hour of day $h \in \{0, \dots, 23\}$ is encoded as $\text{hour}_{\sin} = \sin(2\pi h/24)$ and $\text{hour}_{\cos} = \cos(2\pi h/24)$. For 30-minute intervals, the minutes since midnight $m \in \{0, \dots, 1439\}$ are encoded as $\text{min}_{\sin} = \sin(2\pi m/1440)$ and $\text{min}_{\cos} = \cos(2\pi m/1440)$. This ensures stable, coherent temporal representations across resolutions.

\section{Feature Engineering}

We describe the movement features used for species classification and contrast a minimal baseline representation with an augmented feature set designed to improve learning when data availability varies across species and studies.

\subsection{Baseline Feature Representation}





As a baseline, for each daily trajectory, we extract a minimal 5-dimensional feature vector at each time step $t$: 
$F_{\text{base},t} = [\Delta x_t,\, \Delta y_t,\, \Delta z_t,\, t_{\sin},\, t_{\cos}]$, 
where $(\Delta x_t, \Delta y_t, \Delta z_t)$ represents the spatial displacement on the unit sphere, and $t_{\sin}$ and $t_{\cos}$ refer to the cyclic time-of-day encodings ($\text{hour}$ or $\text{min}$, depending on the resampling resolution) defined in Section III. 

This low-dimensional representation intentionally discards higher-order kinematics (e.g., speed and curvature) to serve as a reference point for evaluating the value of richer motion features.

\subsection{Augmented Movement Features}

To capture higher-level movement dynamics that may be characteristic of different species, we augment the baseline representation with additional kinematic and directional features.

\paragraph{Step Length and Speed}
The step length between consecutive positions is defined as
\begin{equation}
\ell_t = \sqrt{\Delta x_t^2 + \Delta y_t^2 + \Delta z_t^2}.
\end{equation}
Speed is computed as distance traveled per unit time:
\begin{equation}
v_t = \frac{\ell_t}{\Delta t_t},
\end{equation}
where $\Delta t_t$ denotes the time elapsed between consecutive fixes. Speed provides information on movement intensity and is often linked to species-specific locomotion strategies and energy budgets\cite{tucker2018moving}.

\paragraph{Movement Direction (Bearing)}
The instantaneous movement direction is represented by the bearing angle:
\begin{equation}
\theta_t = \arctan2(\Delta y_t, \Delta x_t).
\end{equation}
To avoid discontinuities in $\pm \pi$, the bearing is encoded using its sine and cosine components:
\begin{equation}
\theta_{\sin,t} = \sin(\theta_t), \quad
\theta_{\cos,t} = \cos(\theta_t).
\end{equation}

\paragraph{Turning Angle}
Directional changes between successive steps are captured by the turning angle:
\begin{equation}
\Delta \theta_t = \theta_t - \theta_{t-1},
\end{equation}
wrapped in the interval $[-\pi, \pi]$. As with a bearing, the turning angle is encoded cyclically:
\begin{equation}
\Delta \theta_{\sin,t} = \sin(\Delta \theta_t), \quad
\Delta \theta_{\cos,t} = \cos(\Delta \theta_t).
\end{equation}
Turning behavior reflects path tortuosity and movement persistence, which vary across species and ecological contexts (e.g., foraging vs. traveling)\cite{benhamou2004reliably}.

Together with the baseline features, the augmented representation yields a ten-dimensional feature vector per time step.
\begin{equation}
\begin{aligned}
\{\Delta x_t,\, \Delta y_t,\, \Delta z_t,\, v_t,\, \theta_{\sin,t},\, 
\theta_{\cos,t}, \\
\Delta \theta_{\sin,t},\, \Delta \theta_{\cos,t},\, 
\text{hour}_{\sin},\, \text{hour}_{\cos}\}.
\end{aligned}
\end{equation}
And become like this for 30 min resample:
\begin{equation}
\begin{aligned}
\{\Delta x_t,\, \Delta y_t,\, \Delta z_t,\, v_t,\, \theta_{\sin,t},\, 
\theta_{\cos,t}, \\
\Delta \theta_{\sin,t},\, \Delta \theta_{\cos,t},\, 
\text{min}_{\sin},\, \text{min}_{\cos}\}.
\end{aligned}
\end{equation}

\subsection{Gap-Aware Feature Semantics}

Telemetry data frequently contain missing fixes, leading to irregular time gaps between observations. To avoid inferring artificial movement across such gaps, we adopt a gap-aware feature construction strategy.

Let $\Delta t_t$ denote the time difference between consecutive observations. Movement-derived features (displacements, speed, bearing, and turning) are considered valid only when $\Delta t_t$ is equal to the nominal sampling interval. When this condition is violated, the movement features are marked as undefined rather than imputed:
\begin{equation}
\text{movement\_valid}_t =
\begin{cases}
1, & \text{if } \Delta t_t = \Delta t_{\text{nominal}}, \\
0, & \text{otherwise}.
\end{cases}
\end{equation}

For time steps where $\text{movement\_valid}_t = 0$, all movement-derived features are set to missing values and are handled downstream through masking during sequence modeling. The Time-of-day features remain defined for all timesteps.

This design preserves the semantic distinction between \emph{no movement} and \emph{unknown movement}, preventing the model from learning spurious patterns induced by interpolation or zero-filling. This gap-aware handling is consistent with best practices for temporally structured ecological data and is critical for promising learning under diverse sampling regimes\cite{che2018recurrent}.

\section{Model Architecture}

\subsection{Sequence Modeling Formulation}

We formulate species identification as supervised sequence classification over daily movement trajectories. For each animal-day, we construct a fixed-length sequence
$\mathbf{X} \in \mathbb{R}^{T \times F}$, where $T \in \{24, 48\}$ is the number of timesteps per day (1-hour or 30-minute resampling) and $F \in \{5, 10\}$ is the dimension of the feature (Section IV). Each timestep $\mathbf{x}_t \in \mathbb{R}^F$ contains displacement- and time-derived features; augmented variants also include speed and directional descriptors.

Daily sequences may be shorter than $T$ due to missing fixes. We therefore pad sequences with zeros and provide a binary mask $\mathbf{m} \in \{0,1\}^{T}$ indicating which timesteps are observed:
\begin{equation}
m_t =
\begin{cases}
1, & \text{if timestep } t \text{ is observed},\\
0, & \text{if timestep } t \text{ is padding}.
\end{cases}
\end{equation}
In addition, movement-derived features (e.g. speed, bearing, turning) are treated as undefined across temporal gaps (Section IV-C) and handled via masking / NaN-to-zero conversion at tensor construction time, ensuring that missing movement is not conflated with true zero movement.

\subsection{Transformer Architecture}

Our primary model is a transformer encoder adapted for daily trajectory classification. Each input timestep is first projected to a $d$-dimensional embedding space:
\begin{equation}
\mathbf{z}_t = \mathbf{W}\mathbf{x}_t + \mathbf{b}, \quad \mathbf{W} \in \mathbb{R}^{d \times F}.
\end{equation}
To preserve temporal order, we add sinusoidal positional encodings \cite{vaswani2017attention}:
\begin{equation}
\begin{aligned}
\mathrm{PE}(t,2i)   &= \sin\left(t / 10000^{2i/d}\right), \\
\mathrm{PE}(t,2i+1) &= \cos\left(t / 10000^{2i/d}\right).
\end{aligned}
\end{equation}

and use a learnable classification token (\texttt{[CLS]}) prepended to the sequence. The model applies $L$ pre-norm transformer encoder layers with multi-head self-attention and feed-forward blocks. For an input matrix $\mathbf{Z} \in \mathbb{R}^{T \times d}$, self-attention is calculated as follows:
\begin{equation}
\mathrm{Attn}(\mathbf{Q},\mathbf{K},\mathbf{V}) = 
\mathrm{softmax}\left(\frac{\mathbf{Q}\mathbf{K}^\top}{\sqrt{d_k}}\right)\mathbf{V},
\end{equation}
where $\mathbf{Q}=\mathbf{Z}\mathbf{W}_Q$, $\mathbf{K}=\mathbf{Z}\mathbf{W}_K$, and $\mathbf{V}=\mathbf{Z}\mathbf{W}_V$. Padding positions are excluded using a key padding mask derived from $\mathbf{m}$. The final representation of the \texttt{[CLS]} token is passed through a linear head to produce class logits.

Transformers are well-suited for movement trajectories because they can model long-range temporal dependencies within a day (e.g., multi-hour rest--move cycles) without the recurrence bottlenecks of RNNs, and they can integrate diverse movement cues (displacement, direction, turning, and circadian time) through attention-based aggregation \cite{zerveas2021transformer}.

\subsection{Baseline Models}

To contextualize performance, we compare the transformer against standard sequential architectures commonly used in time-series classification. All baselines were tuned comparably using the exact same validation splits and optimization strategy to ensure a fair comparison:
\begin{itemize}
    \item \textbf{LSTM}: recurrent modeling of temporal dependencies via gated updates (hidden size: 64, 2 layers).
    \item \textbf{1D CNN}: convolutional filters over time capturing local movement motifs. Our architecture utilizes three parallel convolutional branches with kernel sizes of 3, 5, and 7 (64 filters each), followed by Group Normalization (8 groups) and global average pooling.
    \item \textbf{TCN}: dilated causal convolutions enabling wider receptive fields. The network consists of 4 residual blocks (64 hidden channels, kernel size: 3) with exponentially increasing dilation rates ($1, 2, 4, 8$), temporal dropout of 0.2, and masked mean pooling.
\end{itemize}

\begin{table}[!t]
\caption{Models compared in this study.\cite{wang2017time, bai2018empirical}}
\label{tab:models}
\centering
\resizebox{\columnwidth}{!}{%
\begin{tabular}{lccc}
\hline
\textbf{Model} & \textbf{Long Range} & \textbf{Parallelizable} & \textbf{Handle Mask} \\
\hline
LSTM & Moderate & No & Yes \\
1D CNN & Limited & Yes & Yes \\
TCN & Strong (via dilation) & Yes & Yes \\
Transformer & Strong (attention) & Yes & Yes \\
\hline
\end{tabular}%
}
\end{table}

\subsection{Training Objective and Optimization}



Models are trained using a standard cross-entropy loss objective. To mitigate class imbalance without discarding valuable movement data, we apply class-weighted loss, where weights are inversely proportional to class frequencies in the training set. 

We optimize using AdamW\cite{loshchilov2017decoupled} with an initial learning rate of $3\times 10^{-4}$, weight decay of $10^{-4}$, and gradient clipping at a maximum norm of 1.0 and apply dropout\cite{srivastava2014dropout} for regularization. Models were trained with a batch size of 128 for a maximum of 50 epochs. Training was dynamically halted via early stopping (patience of 6 epochs monitoring validation loss) to prevent overfitting. A \texttt{ReduceLROnPlateau} scheduler (factor of 0.5, patience of 2, minimum learning rate $10^{-5}$) dynamically reduced the learning rate when validation improvements stagnated. A fixed random seed strategy was utilized across all experiments to guarantee deterministic data splits and weight initialization.

\subsection{Experimental Setup}

To rigorously test cross-region generalization, we employ a hierarchical, \emph{study-aware} evaluation strategy. For each species, exactly one entire telemetry study is held out exclusively for testing. The remaining data is partitioned into training and validation sets using a strict animal-level split, ensuring all trajectories from a given individual belong entirely to one set. Consequently, hyperparameter tuning relies solely on the validation split without exposure to the test study. This prevents the spatial and temporal leakage inherent in naive random splitting, providing an unbiased test of the model's capacity to learn intrinsic movement patterns \cite{wenger2012assessing}.

Models are trained in a supervised manner using the constructed daily trajectories. To address class imbalance without discarding valuable movement data via aggressive downsampling, we apply a class-weighted cross-entropy loss, where weights are inversely proportional to class frequencies in the training set \cite{buda2018systematic}.

Performance is primarily evaluated using balanced accuracy, which averages recall across classes to effectively handle test set imbalance. We also report the F1 score and the area under the receiver operating characteristic curve (AUC) as secondary metrics \cite{fawcett2006introduction}. Finally, to increase interpretability, we analyze confusion matrices to identify class-wise error asymmetry and species-specific error modes \cite{molnar2020interpretable}.

\section{Results and Performance Analysis}
In this section, we present the experimental results for movement-based species classification. Expanding upon single-species models, we evaluate the performance of seven distinct binary classifiers (One-vs-Rest) encompassing Elephants, Wildebeests, Lions, Buffaloes, Caracals, Baboons, and Zebras. We assess the performance of our Transformer-based architecture against standard sequential baselines, evaluate the impact of kinematic feature enhancement, and analyze the effect of temporal sampling resolution on model accuracy. 

\subsection{Comparison with Sequential Baselines}
We first compare the Transformer against standard sequential baselines—including Long Short-Term Memory (LSTM), 1D Convolutional Neural Networks (1D CNN), and Temporal Convolutional Networks (TCN)—using a 1-hour temporal resolution and our augmented feature set. Table~\ref{tab:baseline_comparison_7species} summarizes the classification performance across all seven species. 

The Transformer architecture consistently outperforms the baselines across the majority of species, demonstrating a superior ability to capture long-range temporal dependencies in daily movement trajectories. Most notably, the Transformer achieves the highest Balanced Accuracy and AUC for Elephants, Wildebeests, Lions, and Buffaloes. While the TCN and 1D CNN models show competitive performance on specific metrics for the Caracal and Zebra, the Transformer maintains highly robust generalization, ensuring stable classification across distinct movement syndromes.

\begin{table}[!t]
\centering
\small
\caption{Performance comparison across species (1-hour, augmented features).}
\label{tab:baseline_comparison_7species}
\resizebox{\columnwidth}{!}{%
\begin{tabular}{l l c c c}
\toprule
Species & Model & Balanced Acc. & F1 & AUC \\
\midrule
\multirow{4}{*}{Elephant}
 & LSTM & 0.68 & 0.69 & 0.78 \\
 & 1D CNN & 0.71 & 0.72 & 0.83 \\
 & TCN & 0.77 & 0.45 & 0.91 \\
 & Transformer & \textbf{0.83} & \textbf{0.84} & \textbf{0.92} \\
\midrule
\multirow{4}{*}{Wildebeest}
 & LSTM & 0.76 & 0.32 & 0.90 \\
 & 1D CNN & 0.78 & 0.39 & 0.95 \\
 & TCN & 0.77 & 0.45 & 0.91 \\
 & Transformer & \textbf{0.85} & \textbf{0.66} & \textbf{0.97} \\
\midrule
\multirow{4}{*}{Lion}
 & LSTM & 0.72 & 0.76 & 0.72 \\
 & 1D CNN & 0.69 & 0.77 & 0.79 \\
 & TCN & 0.79 & 0.81 & 0.88 \\
 & Transformer & \textbf{0.85} & \textbf{0.88} & \textbf{0.91} \\
\midrule
\multirow{4}{*}{Buffalo}
 & LSTM & 0.65 & 0.13 & 0.71 \\
 & 1D CNN & 0.90 & 0.44 & 0.96 \\
 & TCN & 0.89 & 0.51 & 0.96 \\
 & Transformer & \textbf{0.91} & \textbf{0.51} & \textbf{0.98} \\
\midrule
\multirow{4}{*}{Caracal}
 & LSTM & 0.92 & 0.55 & 0.98 \\
 & 1D CNN & 0.93 & 0.64 & 0.98 \\
 & TCN & 0.92 & \textbf{0.85} & 0.98 \\
 & Transformer & \textbf{0.94} & 0.77 & \textbf{0.98} \\
\midrule
\multirow{4}{*}{Baboon}
 & LSTM & 0.77 & 0.71 & 0.99 \\
 & 1D CNN & 0.71 & 0.57 & 0.99 \\
 & TCN & 0.84 & 0.77 & 0.99 \\
 & Transformer & \textbf{0.86} & \textbf{0.81} & \textbf{0.99} \\
\midrule
\multirow{4}{*}{Zebra}
 & LSTM & 0.86 & 0.27 & 0.91 \\
 & 1D CNN & 0.86 & 0.26 & 0.93 \\
 & TCN & \textbf{0.87} & 0.31 & 0.92 \\
 & Transformer & 0.82 & \textbf{0.41} & \textbf{0.93} \\
\bottomrule
\end{tabular}%
}
\end{table}

\subsection{Effect of Feature Augmentation}
To measure the impact of feature engineering, we compared models utilizing a minimal displacement-based feature set (5 features) against those utilizing our augmented feature set (10 features), which includes speed, bearing, and turning descriptors. Both configurations were evaluated using 1-hour resampled trajectories. 

As shown in Table~\ref{tab:feature_augmentation_7species}, incorporating kinematic and directional information yields substantial performance gains. For example, the Balanced Accuracy for the Wildebeest classifier improves from 0.70 to 0.84, and the Lion classifier improves from 0.59 to 0.85. These results confirm that minimal displacement data is often insufficient for distinguishing complex behavioral patterns. Kinematic augmentation is crucial for learning species-specific movement signatures, particularly in data-scarce scenarios. 

\begin{table}[H]
\centering
\small
\caption{Impact of feature augmentation across species (1-hour).}
\label{tab:feature_augmentation_7species}
\resizebox{\columnwidth}{!}{%
\begin{tabular}{l c c c c c c}
\toprule
 & \multicolumn{3}{c}{Minimal (5 Features)} & \multicolumn{3}{c}{Augmented (10 Features)} \\
Species & BalAcc & F1 & AUC & BalAcc & F1 & AUC \\
\midrule
Elephant & 0.58 & 0.78 & 0.69 & \textbf{0.83} & \textbf{0.84} & \textbf{0.92} \\
Wildebeest & 0.70 & 0.21 & 0.75 & \textbf{0.84} & \textbf{0.67} & \textbf{0.97} \\
Lion & 0.59 & 0.53 & 0.67 & \textbf{0.85} & \textbf{0.88} & \textbf{0.91} \\
Buffalo & 0.66 & 0.11 & 0.63 & \textbf{0.91} & \textbf{0.51} & \textbf{0.97} \\
Caracal & 0.67 & 0.18 & 0.68 & \textbf{0.94} & \textbf{0.77} & \textbf{0.97} \\
Baboon & \textbf{0.97} & \textbf{0.93} & 0.99 & 0.92 & 0.88 & \textbf{0.99} \\
Zebra & 0.77 & 0.165 & 0.81 & \textbf{0.82} & \textbf{0.41} & \textbf{0.93} \\
\bottomrule
\end{tabular}%
}
\end{table}

\subsection{Effect of Temporal Resolution}

We also evaluated the sensitivity of the models to temporal sampling by comparing trajectories resampled at 1-hour and 30-minute intervals using the augmented feature set.

Table~\ref{tab:temporal_resolution_7species} shows that temporal resolution affects classification performance across species. Larger herd-based species such as Elephant, Wildebeest, and Buffalo generally achieve higher performance at the 1-hour resolution, which captures stable daily movement patterns while reducing short-term positional noise. In contrast, some species with more dynamic movement behaviors, such as Lions and Caracals, benefit from the finer 30-minute resolution that preserves short-term directional changes.

Overall, however, the 1-hour resampling strategy provides more consistent performance across the multi-study dataset. Because telemetry data in Movebank are collected at heterogeneous sampling frequencies, coarser resampling retains more usable trajectories during preprocessing and improves model robustness across studies.

\begin{table}[!t]
\centering
\small
\caption{Impact of temporal resolution on classification performance (Augmented Features).}
\label{tab:temporal_resolution_7species}
\resizebox{\columnwidth}{!}{%
\begin{tabular}{l c c c c c c}
\toprule
 & \multicolumn{3}{c}{30-Minute} & \multicolumn{3}{c}{1-Hour} \\
Species & BalAcc & F1 & AUC & BalAcc & F1 & AUC \\
\midrule
Elephant & 0.77 & 0.72 & 0.84 & \textbf{0.83} & \textbf{0.84} & \textbf{0.92} \\
Wildebeest & 0.76 & 0.61 & 0.72 & \textbf{0.84} & \textbf{0.66} & \textbf{0.97} \\
Lion & \textbf{0.90} & \textbf{0.96} & \textbf{0.98} & 0.8555 & 0.877 & 0.91 \\
Buffalo & 0.87 & 0.46 & 0.95 & \textbf{0.91} & \textbf{0.51} & \textbf{0.97} \\
Caracal & \textbf{0.96} & \textbf{0.89} & \textbf{0.98} & 0.94 & 0.77 & 0.98 \\
Baboon & 0.77 & 0.71 & 0.99 & \textbf{0.86} & \textbf{0.81} & \textbf{0.99} \\
Zebra & \textbf{0.93} & \textbf{0.53} & \textbf{0.98} & 0.82 & 0.41 & 0.93 \\
\bottomrule
\end{tabular}%
}
\end{table}

\subsection{Confusion Matrices and Final Performance}
\begin{figure}[t]
\centering
\scriptsize

\begin{subfigure}{0.2\textwidth}
\includegraphics[width=\linewidth]{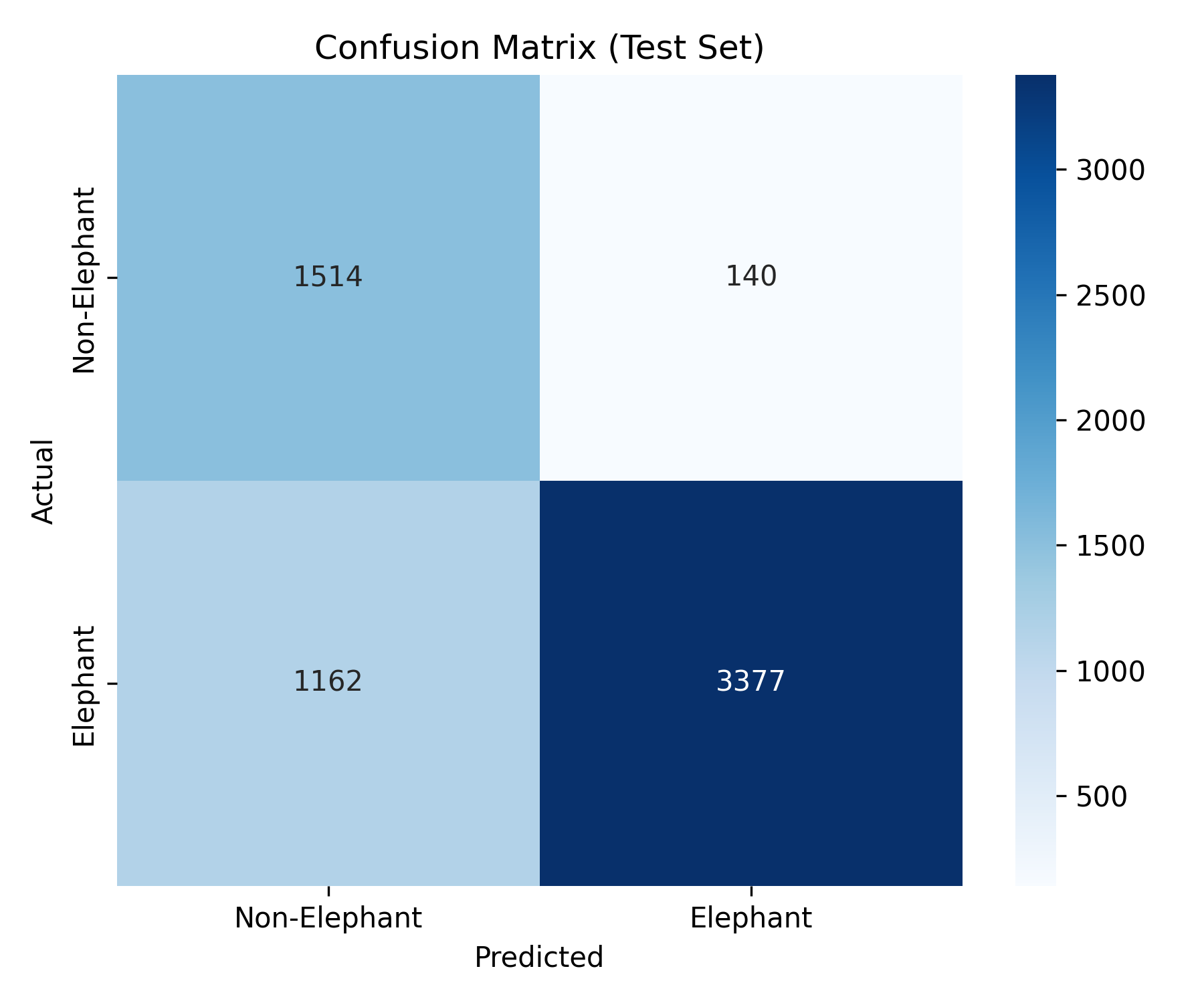}
\caption{Elephant (1h)}
\end{subfigure}
\hfill
\begin{subfigure}{0.2\textwidth}
\includegraphics[width=\linewidth]{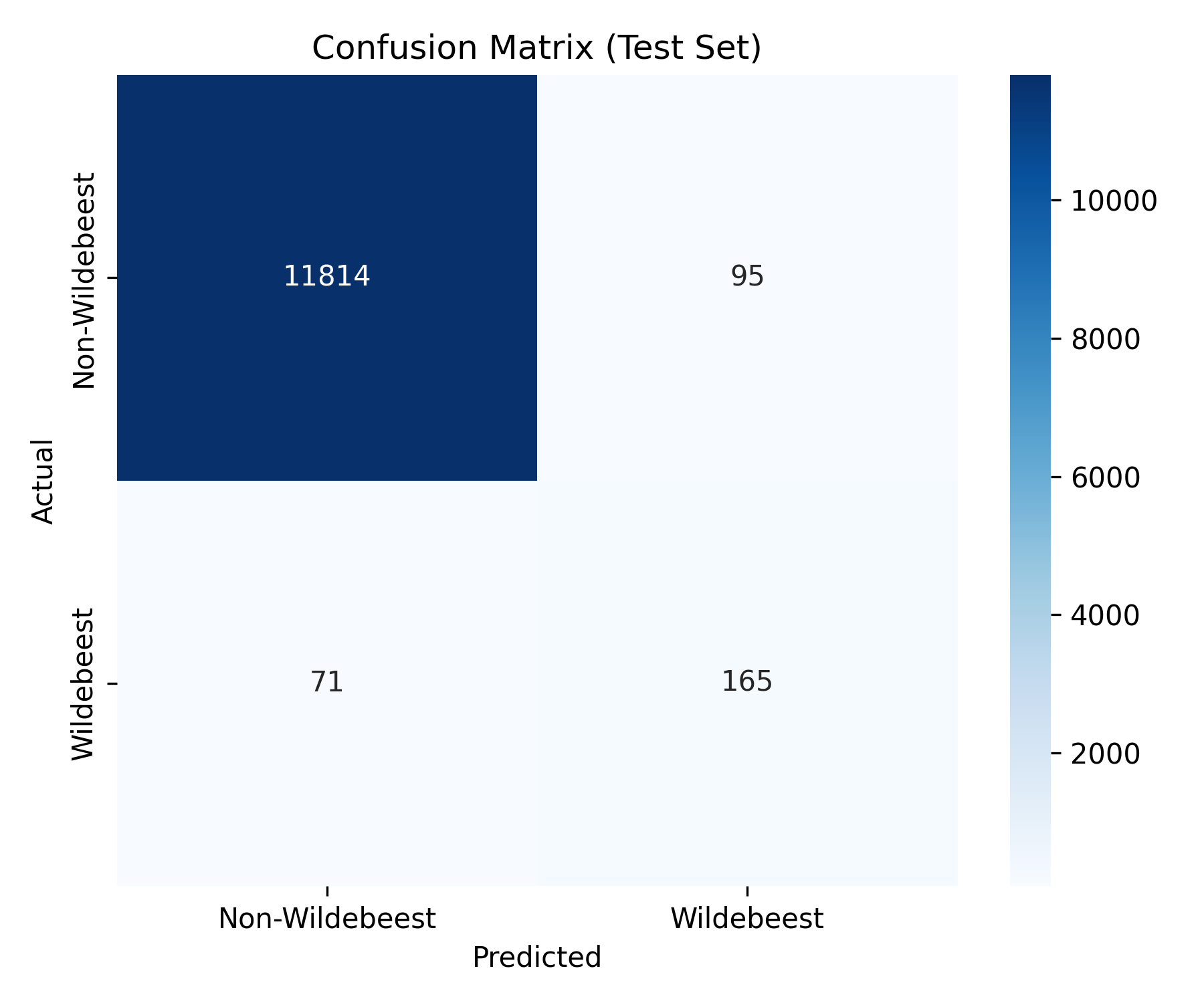}
\caption{Wildebeest (1h)}
\end{subfigure}

\vspace{0.1cm}

\begin{subfigure}{0.2\textwidth}
\includegraphics[width=\linewidth]{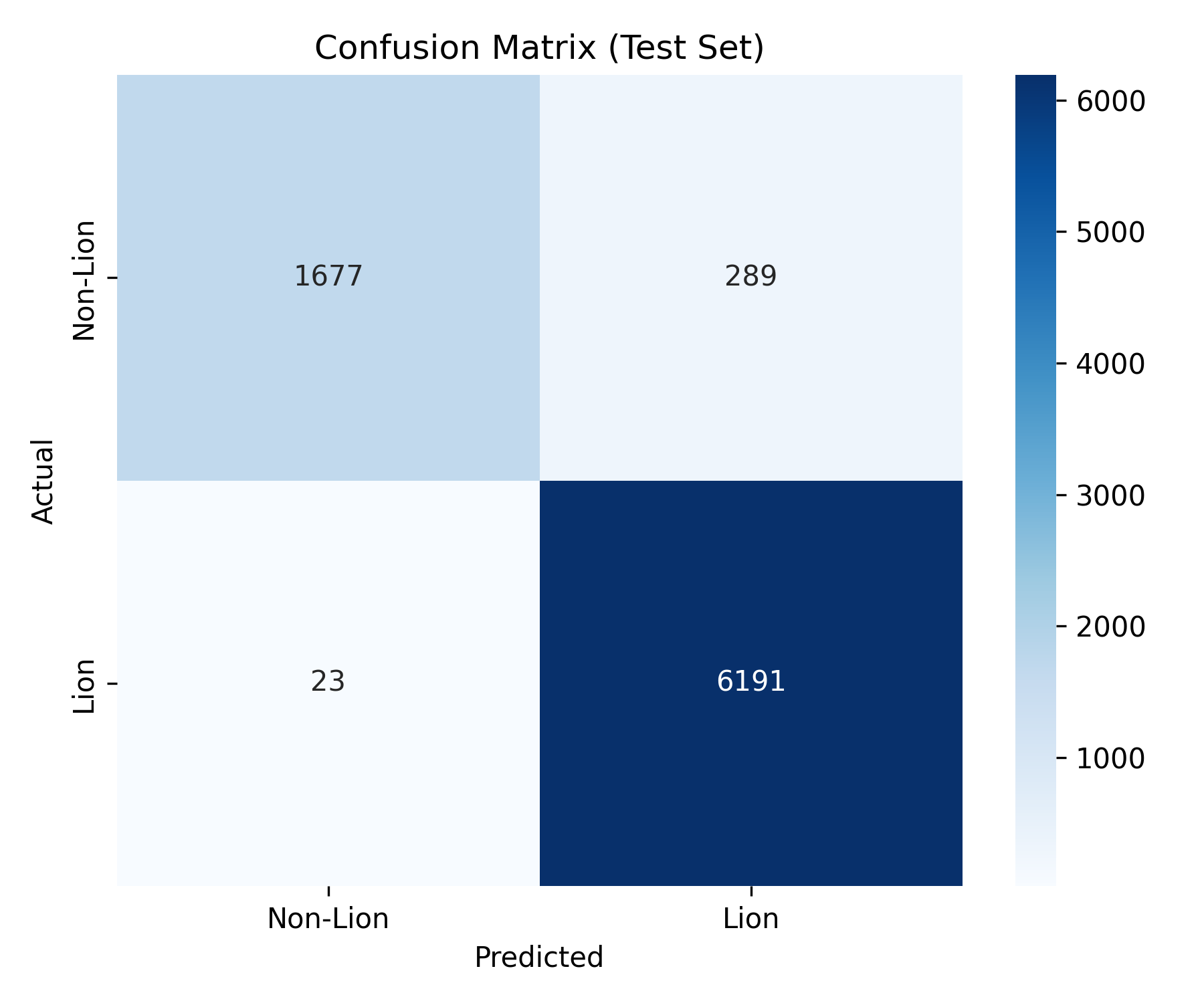}
\caption{Lion (30m)}
\end{subfigure}
\hfill
\begin{subfigure}{0.2\textwidth}
\includegraphics[width=\linewidth]{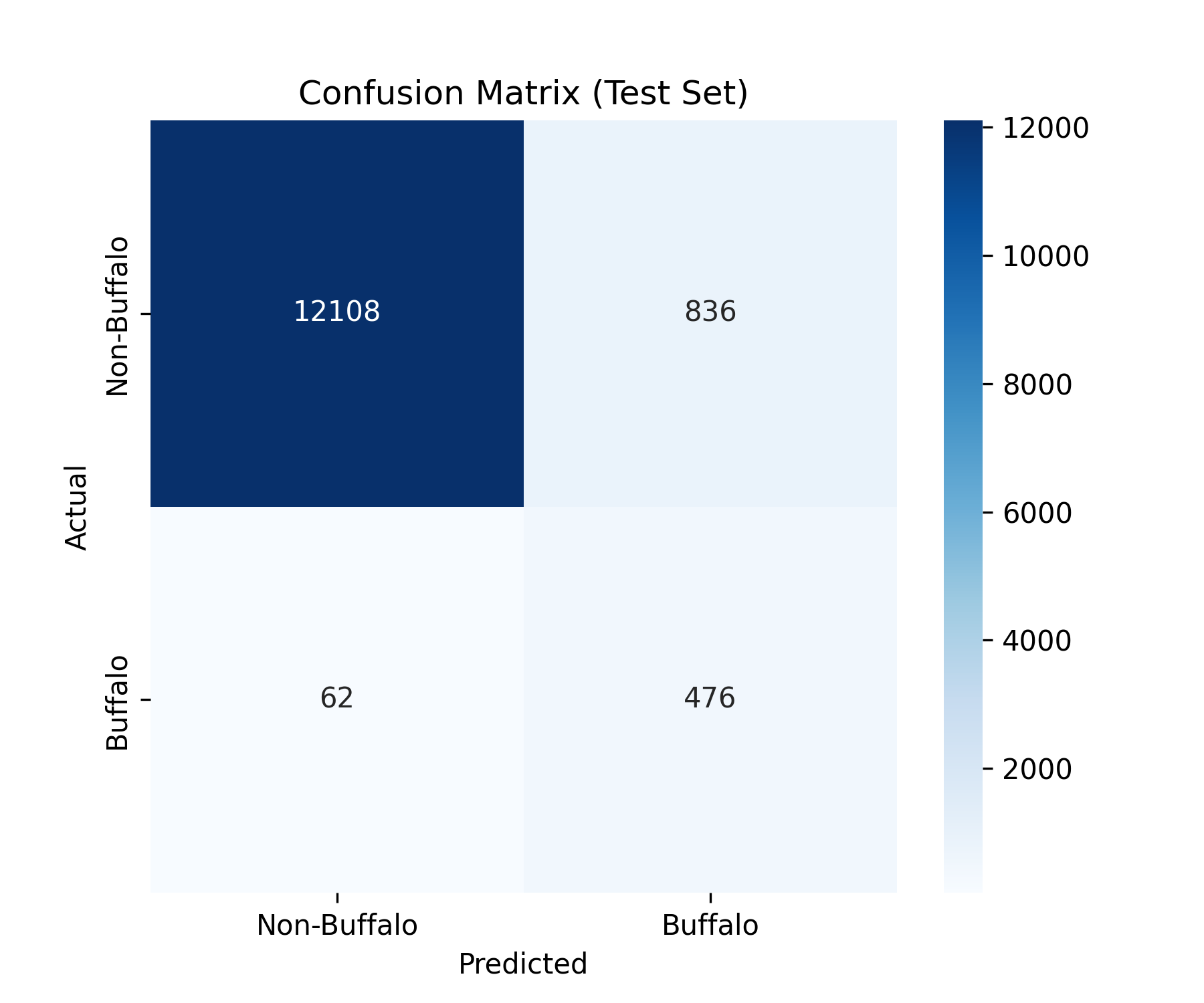}
\caption{Buffalo (1h)}
\end{subfigure}

\vspace{0.1cm}

\begin{subfigure}{0.2\textwidth}
\includegraphics[width=\linewidth]{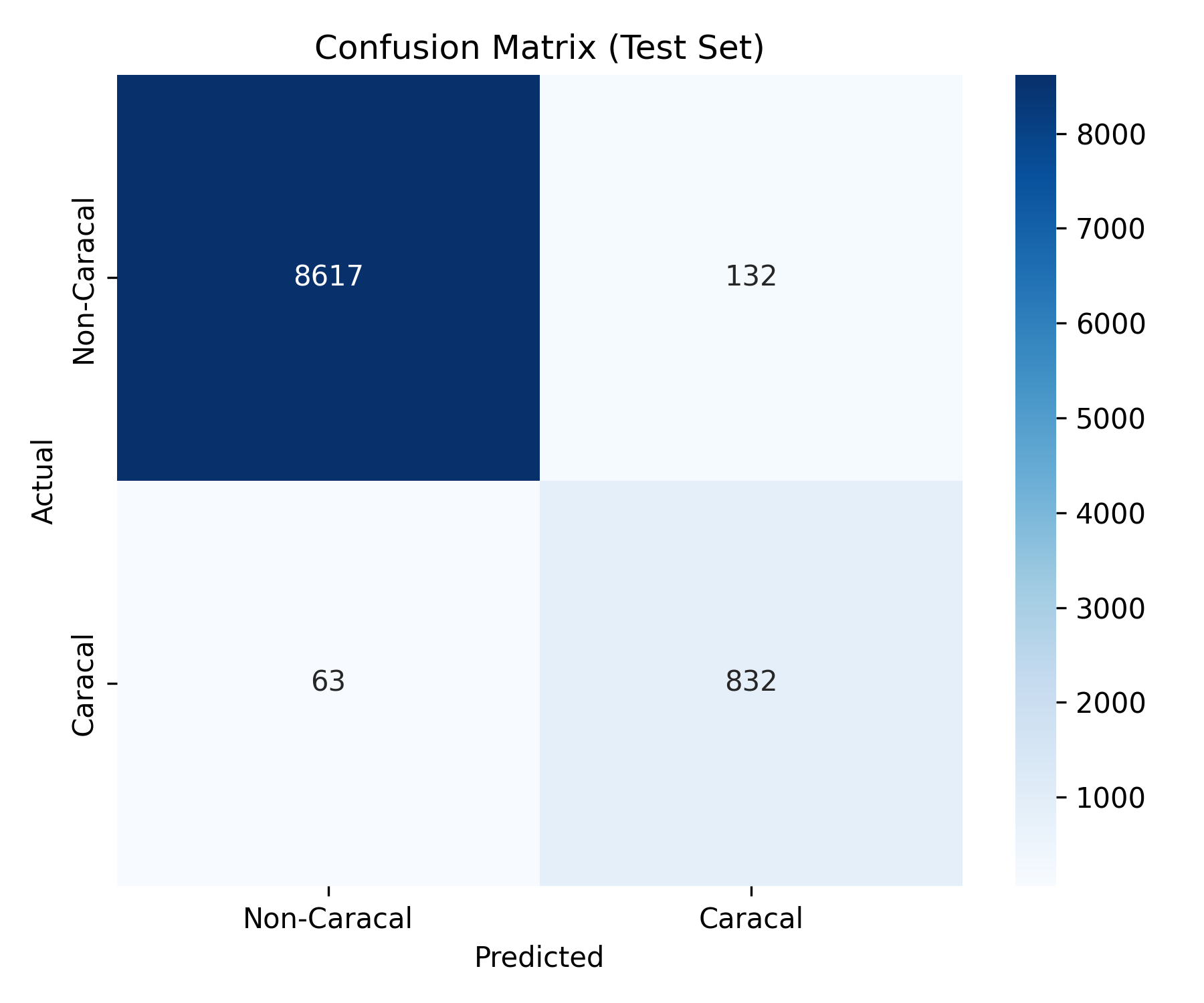}
\caption{Caracal (30m)}
\end{subfigure}
\hfill
\begin{subfigure}{0.2\textwidth}
\includegraphics[width=\linewidth]{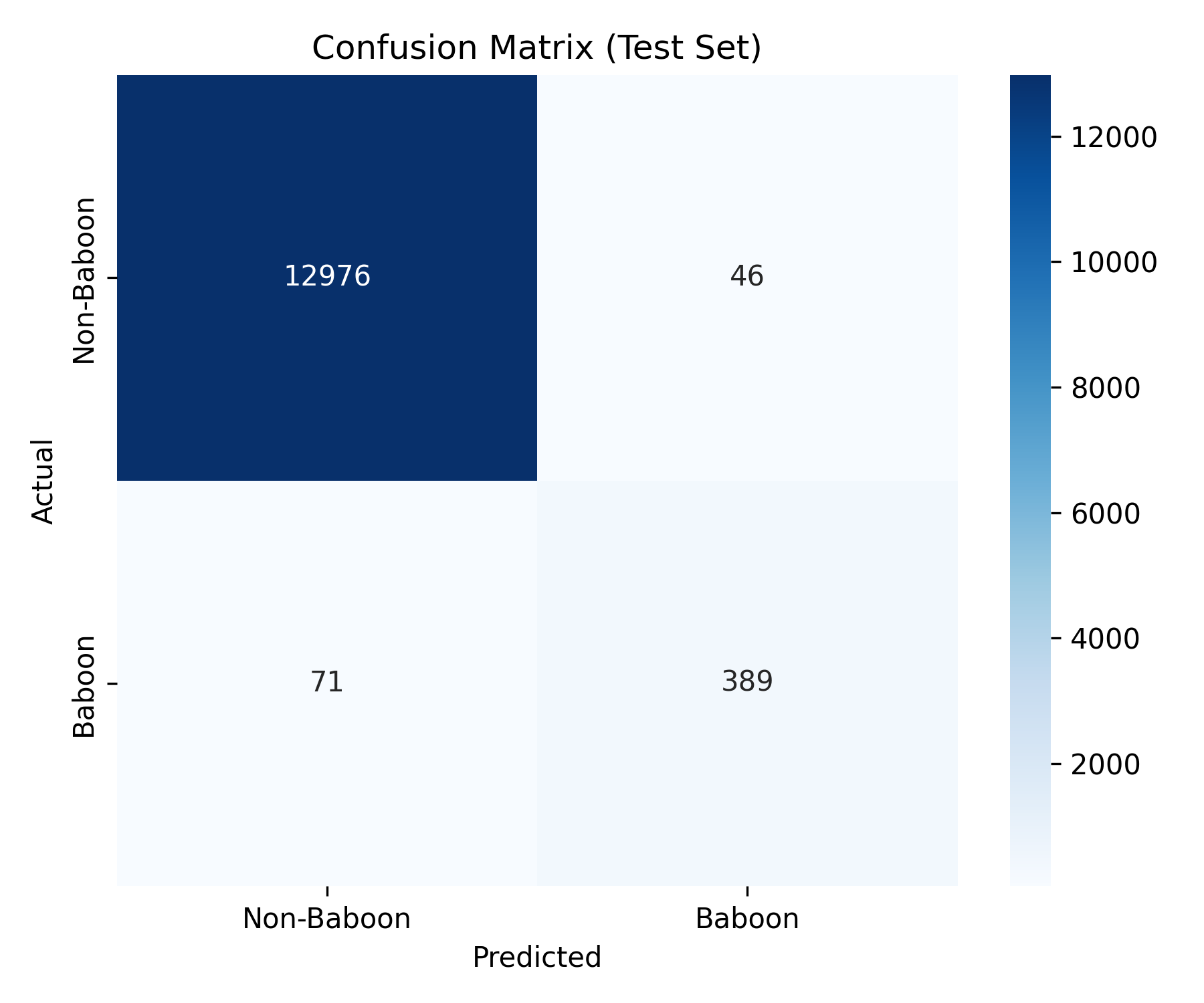}
\caption{Baboon (1h)}
\end{subfigure}

\vspace{0.1cm}

\begin{subfigure}{0.2\textwidth}
\includegraphics[width=\linewidth]{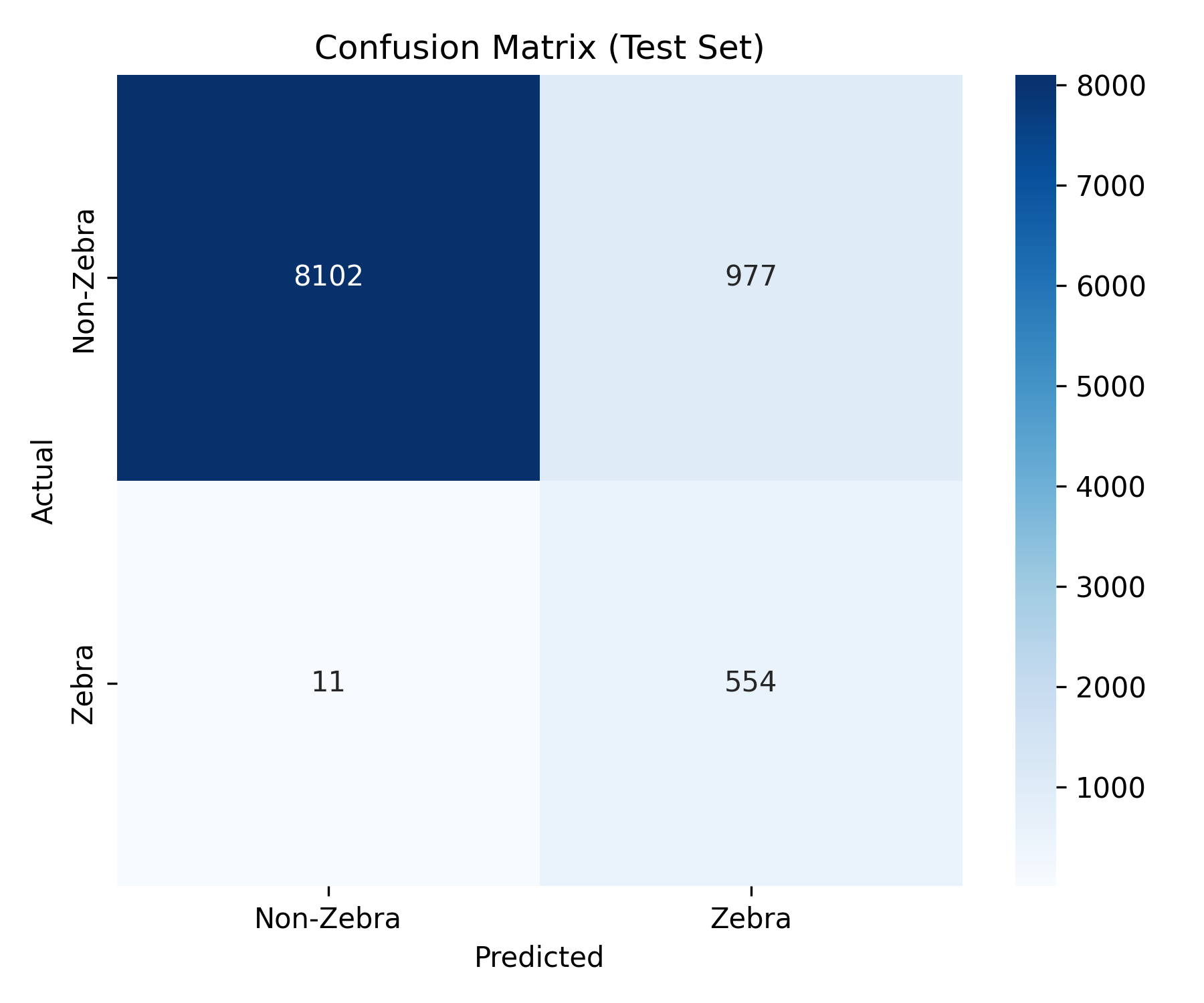}
\caption{Zebra (30m)}
\end{subfigure}

\caption{Confusion matrices for the best-performing Transformer configurations across species.}
\label{fig:confusion_matrices_all}
\end{figure}

Figure~\ref{fig:confusion_matrices_all} visualizes the confusion matrices for the best-performing Transformer configurations for each species (utilizing their optimal temporal resolutions). The matrices demonstrate high recall for the target species while maintaining strong specificity against the combined negative classes. This high discrimination capability highlights the suitability of the Transformer architecture for deployment in real-world wildlife monitoring and early-warning systems.

\section{Conclusion}

This paper investigated whether the identity of wildlife species can be inferred from movement trajectories alone, using daily sequences of GPS coordinates collected in multiple
regions and telemetry campaigns. Using large-scale African wildlife data from Movebank, we showed that transformer-based sequence models outperform established sequential baselines model for wildlife species classification under cross-study evaluation. In the elephant binary classifier, the transformer achieved a balanced accuracy of 0.83 and an AUC of 0.92,with significant gains over the models based on LSTM, CNN, and TCN.

We further demonstrated that augmenting simple displacement features with movement descriptors capturing speed, direction, and turning behavior markedly improves performance, and that temporal resolution plays a significant role, where a one-hour temporal resampling rate generally provides more promising performance than the 30-minute resampling rate under data with variable temporal resolution conditions.Together, these results indicate that daily movement trajectories could encode discriminative movement patterns that remain informative across different regions and telemetry studies, and that attention-based models are well-suited to capturing them under realistic and cross-region data constraints.

\section{Future Work}

Several directions emerge from this study. 
\begin{itemize}

\item \textbf{Improved augmentation of movement feature sets:} Building upon the existing movement feature sets to ascertain whether more movement descriptors improve the performance of the models or whether the performance becomes saturated.

\item \textbf{Multiday trajectory modeling:} Building upon the existing one-day movement trajectory modeling to include movement continuity over two or three consecutive days to help the models discover hidden constraints that may not be evident in one-day trajectory modeling.

\item \textbf{Multiple species classification:} Building upon binary classification to include multiple species classification of all considered species.

\item \textbf{Wider ecological coverage:} Expanding the coverage of species and regions as open-access telemetry data become available.


\end{itemize}
\section{Acknowledgement}
This publication was developed as part of the Center for Inclusive Digital Transformation of Africa (CIDTA), and, the Afretec Network which is managed by Carnegie Mellon University Africa and receives financial support from the Mastercard Foundation. 
The views expressed in this document are
solely those of authors and do not necessarily reflect those of the Carnegie Mellon University or
the Mastercard Foundation.

\bibliographystyle{IEEEtran}
\bibliography{Reference}

\end{document}